\documentclass[10pt,twocolumn,letterpaper]{article}

\usepackage{cvpr}  
\usepackage{graphicx}
\usepackage{amsmath,amssymb,graphicx,booktabs}
\usepackage{booktabs}  
\usepackage{multirow}  
\usepackage{graphicx}  
\usepackage{array}     
\usepackage{subcaption}

\definecolor{cvprblue}{rgb}{0.21,0.49,0.74}
\usepackage[pagebackref,breaklinks,colorlinks,allcolors=cvprblue]{hyperref}

\def\paperID{*****} 
\def\confName{CVPR}
\def\confYear{2026}

\title{TriTS: Time Series Forecasting from a Multimodal Perspective}
\author{
    Xiang Ao\thanks{Accepted by the A2A-MML Workshop in conjunction with IEEE/CVF CVPR 2026.} \\
    School of Software Engineering, Beijing Jiaotong University \\
    Beijing 100044, China \\
    {\tt\small ao.xiang.axel@outlook.com}
}
\begin{document}
\maketitle
\begin{abstract}
Time series forecasting plays a pivotal role in critical sectors such as finance, energy, transportation, and meteorology. However, Long-term Time Series Forecasting (LTSF) remains a significant challenge because real-world signals contain highly entangled temporal dynamics that are difficult to fully capture from a purely 1D perspective. To break this representation bottleneck, we propose TriTS, a novel cross-modal disentanglement framework that projects 1D time series into orthogonal time, frequency, and 2D-vision spaces.To seamlessly bridge the 1D-to-2D modality gap without the prohibitive $O(N^2)$ computational overhead of Vision Transformers (ViTs), we introduce a Period-Aware Reshaping strategy and incorporate Visual Mamba (Vim). This approach efficiently models cross-period dependencies as global visual textures while maintaining linear computational complexity. Complementing this, we design a Multi-Resolution Wavelet Mixing (MR-WM) module for the frequency modality, which explicitly decouples non-stationary signals into trend and noise components to achieve fine-grained time-frequency localization. Finally, a streaming linear branch is retained in the time domain to anchor numerical stability. By dynamically fusing these three complementary representations, TriTS effectively adapts to diverse data contexts. Extensive experiments across multiple benchmark datasets demonstrate that TriTS achieves state-of-the-art (SOTA) performance, fundamentally outperforming existing vision-based forecasters by drastically reducing both parameter count and inference latency.
\end{abstract}    
\section{Introduction}
\label{sec:intro}

Time series forecasting plays an indispensable role in energy load scheduling, meteorological disaster early warning, financial transactions, and other critical fields \cite{energe,weather,finance,traffic,industrial}. As prediction windows expand from short-term to long-term, models are required to infer future trends by capturing complex long-range dependencies from massive historical data. However, real-world time series are inherently complex, featuring highly entangled components such as global trends, complex seasonality, and non-stationary local mutations.In recent years, the forecasting paradigm has advanced from early Recurrent Neural Networks (RNNs) \cite{rnn,lstm} and Transformer-based architectures \cite{attention,autoformer,patchtst} towards a broader cross-modal representation perspective. Recognizing the representation bottleneck of analyzing signals solely in the 1D time domain, researchers have sought to project time series into alternative modalities. Frequency-domain methods \cite{fedformer,filternet} isolate periodic patterns, while pioneering vision-based approaches (e.g., VisionTS \cite{visionts}, VisionTS++ \cite{visionts++}) have demonstrated that folding 1D signals into 2D "temporal images" can successfully leverage the powerful structural perception capabilities of vision foundation models, achieving remarkable cross-domain transferability.Despite these advances, existing cross-modal bridging strategies still suffer from severe computational and representational limitations. First, in the vision domain, current methods predominantly rely on standard Vision Transformer (ViT) architectures. The self-attention mechanism inherently dictates a quadratic computational complexity ($O(N^2)$) with respect to the sequence length \cite{manba}, leading to prohibitive memory consumption and inference latency when processing ultra-long historical windows. Second, in the frequency domain, methods relying on the Fast Fourier Transform (FFT) utilize global basis functions, making it exceedingly difficult to accurately localize the transient mutations and non-stationary shifts prevalent in real-world data. Consequently, current models often either sacrifice computational efficiency for global context or lose fine-grained temporal resolution.To overcome the representation bottlenecks of single modalities and dismantle the computational barriers of cross-modal forecasting, this paper proposes TriTS, a unified cross-modal disentanglement framework that seamlessly synergizes time, frequency, and vision modalities. Rather than simply concatenating different models, TriTS fundamentally rethinks time series as a multimodal entity, routing specific signal characteristics to their most natural processing domains.Specifically, to bridge the 1D-to-2D modality gap without the heavy burden of ViTs, TriTS introduces a novel vision pathway integrating Period-Aware Reshaping with Visual Mamba (Vim) \cite{vim}. By utilizing the selective scanning mechanism of the bidirectional State Space Model (SSM), the Vim-Encoder treats periodic variations as 2D visual textures, efficiently capturing cross-period global dependencies with linear ($O(N)$) computational complexity. For the frequency modality, we propose a Multi-Resolution Wavelet Mixing (MR-WM) module to replace traditional global FFTs. Through multi-level wavelet decomposition, this module provides physical priors by explicitly decoupling non-stationary signals into trend and detail components, achieving precise time-frequency localization. Finally, to mitigate distribution shifts and anchor numerical stability, we deploy a streaming Exponential Moving Average (EMA) linear model \cite{xpatch} in the time domain. These three orthogonal, yet complementary, representations are dynamically aggregated via a scale-aware gating mechanism, adapting robustly to diverse datasets.Extensive experiments on multiple benchmark datasets demonstrate that TriTS achieves state-of-the-art performance. Crucially, compared to existing vision-based forecasters, TriTS drastically reduces parameter count and inference latency, redefining the efficiency baseline for multimodal time series analysis.The main contributions of this paper are summarized as follows:

\begin{enumerate}

\item We propose TriTS, a novel cross-modal disentanglement framework that projects 1D time series into time, frequency, and 2D-vision spaces. By dynamically fusing these orthogonal perspectives, TriTS effectively resolves the information loss and representation bottlenecks inherent in single-modality methods.

\item We pioneer the integration of Visual Mamba (Vim) into time series forecasting via a 1D-to-2D modality bridging strategy. By leveraging bidirectional State Space Models, we break the $O(N^2)$ computational bottleneck of traditional vision-based forecasters, enabling efficient global texture capture in ultra-long sequences.

\item We introduce a frequency-domain encoder based on Multi-Resolution Wavelet Mixing (MR-WM) to address the limitations of FFTs in handling non-stationary mutations. This module achieves hierarchical decoupling of trends and noise, significantly enhancing the accurate localization of local anomalies and complex seasonality.
\end{enumerate}
\section{Related work}

\subsection{Time domain}
Deep learning-based time series forecasting has evolved from early Recurrent Neural Networks (RNNs)\cite{rnn,lstm} to the currently dominant Transformer architectures. Models such as LogTrans\cite{logtrans}, Autoformer\cite{autoformer}, and Fedformer\cite{fedformer} introduced various attention mechanisms to capture long-range dependencies. Recently, PatchTST\cite{patchtst} and iTransformer\cite{itransformer} significantly improved performance by adopting patch-based processing and inverted channel attention, respectively. However, the quadratic computational complexity of the self-attention mechanism remains a bottleneck for handling ultra-long sequences. In response, simple yet effective linear models like DLinear\cite{dlinear} and TimeMixer\cite{timemixer} have emerged, challenging the necessity of complex architectures. While linear models are efficient, they often lack the capacity to capture complex non-linear dynamics and global semantic contexts compared to deep models. Our work integrates a streaming linear branch to preserve the robustness of linear trends while leveraging other modalities for complex feature extraction.

\subsection{Frequency domain}
Frequency domain analysis offers a unique perspective for capturing periodic patterns in time series. Several studies have attempted to combine deep learning with frequency transforms\cite{fedformer,fdnet,filternet}. However, FFT-based methods and global filters rely on global basis functions, making them less effective at capturing local non-stationary variations and transient anomalies. In contrast, wavelet-based methods such as WPMixer\cite{wpmixer} and FDNet\cite{fdnet} provide superior time-frequency localization capabilities. Specifically, WPMixer utilizes wavelet packet decomposition to disentangle multi-scale features, enabling the model to capture fine-grained details that are often missed by global frequency transforms. Building on this, TriTS adopts a multi-resolution wavelet mixing strategy to explicitly disentangle trend and noise components, further addressing the limitations of global frequency operations in non-stationary environments.

\subsection{Vision domain}
Inspired by the immense success of computer vision\cite{vit,resnet,yolo}, recent research explores transforming time series into image-like formats to leverage powerful vision backbones. TimesNet\cite{timesnet} transforms 1D time series into 2D tensors based on multiple periods, utilizing 2D convolutional kernels to capture intra-period and inter-period variations, effectively treating temporal variations as visual patterns. Taking this cross-modal transfer further,VisionTS\cite{visionts} and VisionTS++\cite{visionts++} reformulate time series forecasting as an image reconstruction task. They convert time series into 2D images and leverage pre-trained visual Masked Autoencoders (MAE) to capture high-level semantic dependencies, achieving state-of-the-art performance in zero-shot and few-shot scenarios. However, these standard vision backbones (like ViT\cite{vit,mae} used in VisionTS) inherit the heavy computational burden of the self-attention mechanism, scaling quadratically with sequence length. Recently, State Space Models (SSMs), particularly Mamba\cite{manba}, have gained attention for their ability to model long sequences with linear complexity ($O(N)$). Visual Mamba (Vim)\cite{vim} extends this to the vision domain, offering a more efficient alternative to ViT. TriTS introduces Visual Mamba into the multimodal time series forecasting framework, successfully combining the global receptive field of vision models with the efficiency required for long-term forecasting.
\section{Methodology}

\subsection{Problem Formulation and Overview}
The Long-Term Time Series Forecasting (LTSF) task aims to infer future temporal evolution based on historical observations. Formally, given a multivariate historical time series $\mathbf{X} = \{\mathbf{x}_1, \dots, \mathbf{x}_L\} \in \mathbb{R}^{L \times C}$, where $L$ denotes the lookback window length and $C$ represents the number of channels, our goal is to predict the future sequence $\hat{\mathbf{Y}} = \{\hat{\mathbf{x}}_{L+1}, \dots, \hat{\mathbf{x}}_{L+T}\} \in \mathbb{R}^{T \times C}$.

\begin{figure*}[htbp]
    \centering
    \includegraphics[width=\linewidth]{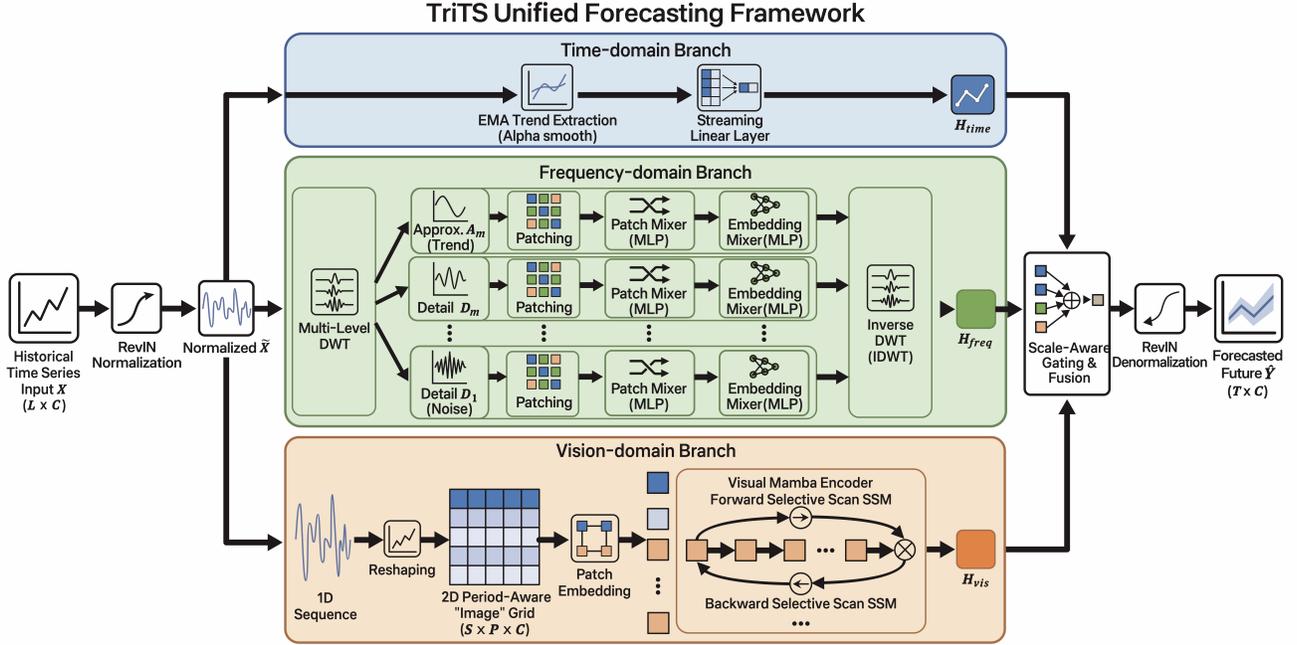}
    \caption{Schematic illustration of the proposed TriTS framework. It integrates three modalities to disentangle trends and noise. Key components include the Visual Mamba (Vim) for cross-period dependency modeling and the Multi-Resolution Wavelet Mixing (MR-WM) for fine-grained time-frequency localization.}
    \label{fig:arch}
\end{figure*}

Real-world time series data exhibit high complexity, characterized by intertwined local variations, global trends, and intricate seasonality. Single-modality models often struggle to balance these multifaceted characteristics. To overcome the representation bottleneck inherent in single-perspective approaches, we propose the \textbf{TriTS} framework. As illustrated in Figure \ref{fig:arch}, TriTS extracts features from the time, frequency, and vision domains through three complementary parallel branches. Specifically, the time-domain branch employs a streaming linear model to maintain numerical integrity; the frequency-domain branch utilizes multi-resolution wavelet mixing to decouple trends from noise; and the vision branch leverages Visual Mamba to efficiently capture global textures. To mitigate the distributional shift caused by non-stationarity, the input for all branches is the sequence $\tilde{\mathbf{X}}$, normalized via RevIN.

\subsection{Time-domain Branch}
In the time-domain branch, we aim to construct a robust "linear anchor" to capture non-stationary trend terms. Although simple Linear Layers are effective, they are overly sensitive to high-frequency noise within the input sequence. While existing methods employ Simple Moving Average (SMA) for trend extraction, SMA often results in excessive information smoothing and necessitates padding at sequence boundaries, thereby introducing boundary bias.

Inspired by xPatch\cite{xpatch}, we adopt Exponential Moving Average (EMA) as the trend extraction operator. By assigning weights that decay exponentially over time, EMA responds more sensitively to recent temporal changes while maintaining consistent sequence length without requiring padding operations.

Given the normalized input sequence $\tilde{\mathbf{X}}$, we first decompose the trend term $\mathbf{X}_{\mathrm{trend}}$ using EMA. For the input $\tilde{\mathbf{x}}_t$ at time $t$, its trend is updated as follows:
\begin{equation}
    \mathbf{X}_{\mathrm{trend}, t} = \alpha \tilde{\mathbf{X}}_t + (1-\alpha) \mathbf{X}_{\mathrm{trend}, t-1}
\end{equation}
where $\alpha \in (0, 1)$ is the smoothing coefficient. Subsequently, to prevent the activation function from disrupting the linear relationship, we feed the extracted trend term into a streaming linear layer to obtain the time-domain branch output $\mathbf{H}_{\mathrm{time}}$:
\begin{equation}
    \mathbf{H}_{\mathrm{time}} = \mathcal{L}_{\mathrm{stream}}(\mathbf{X}_{\mathrm{trend}}) = \mathbf{X}_{\mathrm{trend}}\mathbf{W}_{\mathrm{time}} + \mathbf{b}_{\mathrm{time}}
\end{equation}
where $\mathbf{W}$ is the weight of the linear model and $\mathbf{b}$ is the bias term. This design combines the advantage of EMA in capturing dynamic trends and provides an unbiased low-frequency benchmark for the entire TriTS framework.

\subsection{Frequency-domain Branch}
To overcome the limitations of FFT in modeling non-stationary signals, we adopt a wavelet-based multi-resolution branch. Inspired by the architectural design of WPMixer\cite{wpmixer}, we construct independent Resolution Branches for each frequency component and introduce a dual mixing mechanism for Patch-Embedding.

\subsubsection{Multi-Level Decomposition \& Independent Branches}
To achieve physically meaningful frequency-domain decoupling, we decompose the input sequence $\tilde{\mathbf{X}}$ using multi-level Discrete Wavelet Transform (DWT) based on the Mallat algorithm. For each decomposition level $i$ ($1 \le i \le m$), the approximation coefficients from the previous level generate the current approximation coefficients $\mathbf{X}_{A_i}$ and detail coefficients $\mathbf{X}_{D_i}$ via high-low pass filters and downsampling operations. To eliminate information redundancy, we strictly retain only the top-level approximation coefficients $\mathbf{X}_{A_m}$, representing long-term macro trends, and the set of detail coefficients $\{ \mathbf{X}_{D_1}, \dots, \mathbf{X}_{D_m} \}$, which captures local variations at each level:
\begin{equation}
    \mathbf{X}_{A_i}, \mathbf{X}_{D_i} = \text{DWT}(\mathbf{X}_{A_{i-1}})
\end{equation}

\begin{equation}
    \quad \mathcal{W}(\tilde{\mathbf{X}}) = \{ \mathbf{X}_{A_m}, \mathbf{X}_{D_m}, \dots, \mathbf{X}_{D_1} \}
\end{equation}

Unlike traditional methods that process frequency-domain features in a mixed manner, we construct completely independent resolution branches for these $m+1$ components. Each branch possesses an independent parameter space, including dedicated normalization (RevIN), Patching, and mixer modules. This resolution isolation strategy rigorously prevents high-frequency noise from contaminating low-frequency trends during feature extraction, ensuring that different frequency components are accurately modeled at their respective physical scales. 

Crucially, this architecture inherently resolves the issue of periodicity harmonics. By distributing fundamental frequencies and their higher-order harmonics into distinct wavelet sub-bands ($\mathbf{X}_{D_i}$), our independent branch design explicitly decouples them. This effectively mitigates the harmonic interference and frequency leakage commonly encountered in global FFT operations. Finally, the independent predictions from each branch are structurally reconstructed via Inverse Discrete Wavelet Transform (IDWT) to generate the frequency-domain branch output $\mathbf{H}_{\mathrm{freq}}$.

\subsubsection{Dual-Stage Mixing: Patch \& Embedding}
Within each resolution branch, to efficiently capture long-range dependencies and cross-variable correlations, we employ a Dual-Stage Mixer module. First, we segment the wavelet coefficient sequence into Patches. The Patch Mixer operates along the Patch dimension to capture local temporal context, aggregating information via transposition operations and MLPs. Subsequently, the Embedding Mixer operates along the channel (Embedding) dimension to extract global semantic features through a shared MLP.

Formally, for the representation $\mathbf{Z}_k$ of the $k$-th frequency component, the calculation process is defined as:
\begin{equation}
    \mathbf{Z}_k^{\mathrm{patch}} = \text{MLP}_{\mathrm{patch}}(\text{Permute}(\mathbf{Z}_k^{\mathrm{in}}))
\end{equation}
\begin{equation}
    \mathbf{H}_{f, k} = \text{MLP}_{\mathrm{emb}}(\text{Permute}(\mathbf{Z}_k^{\mathrm{patch}})) + \mathbf{Z}_k^{\mathrm{patch}}
\end{equation}

This design enables the model to leverage both local details in the time domain and global structures in the frequency domain, achieving correlation modeling with $O(L)$ complexity.

\subsection{Vision-domain Branch}
To circumvent the $O(L^2)$ computational bottleneck inherent in ViT\cite{vit} used by standard VisionTS\cite{visionts} when processing long sequences, this branch introduces Vim\cite{vim}. Leveraging the selective scanning mechanism of the State Space Model (SSM), Vim efficiently captures global texture features of time series—analogous to image processing—while maintaining linear computational complexity. Specifically, the dominant period $P$ is determined via the autocorrelation function, which effectively identifies long-term stable periodicities in time series forecasting. This period-aware perspective allows the model to reshape 1D sequences into 2D temporal images, capturing both intra-period variations and inter-period evolutions.

\subsubsection{Period-Aware Reshaping \& Patching}
Time series often exhibit strong periodicity. To adapt vision models for 1D signal processing, we first fold the normalized sequence $\tilde{\mathbf{X}} \in \mathbb{R}^{L \times C}$ based on the dominant period $P$, reshaping it into a "temporal image" $\mathbf{I} \in \mathbb{R}^{S \times P \times C}$, where $S = L/P$. This transformation maps intra-period high-frequency variations to row textures and inter-period evolution to column structures.
Subsequently, we partition $\mathbf{I}$ into non-overlapping Patch sequences $\mathbf{x}_p \in \mathbb{R}^{N \times (P^2 \cdot C)}$ and project them into the latent space $\mathbf{H}_0$ to prepare for the Vim encoder.

\subsubsection{Bidirectional Selective State Space Modeling}
The core innovation of Vim lies in replacing the traditional self-attention mechanism with a bidirectional State Space Model (Bi-SSM). For the input sequence $x(t)$, the continuous-time SSM establishes a mapping through the hidden state $h(t)$:
\begin{equation}
    h'(t) = \mathbf{A}h(t) + \mathbf{B}x(t), \quad y(t) = \mathbf{C}h(t)
\end{equation}

To perform calculations on discrete time steps, we introduce the time-scale parameter $\mathbf{\Delta}$ and discretize the parameters $(\mathbf{A}, \mathbf{B})$ into $(\overline{\mathbf{A}}, \overline{\mathbf{B}})$ using Zero-Order Hold (ZOH):
\begin{equation}
    \overline{\mathbf{A}} = \exp(\mathbf{\Delta} \mathbf{A}), \quad \overline{\mathbf{B}} = (\mathbf{\Delta} \mathbf{A})^{-1}(\exp(\mathbf{\Delta} \mathbf{A}) - \mathbf{I}) \cdot \mathbf{\Delta} \mathbf{B}
\end{equation}

Unlike traditional SSMs, Mamba introduces a Selectivity mechanism, where $\mathbf{B}, \mathbf{C}, \mathbf{\Delta}$ are all dynamically generated by the input. This enables the model to adaptively "memorize" or "forget" historical information based on the context.

Since visual perception lacks strict causality (i.e., pixel dependencies are bidirectional), unidirectional scanning results in information asymmetry. Therefore, our Vim encoder employs Bidirectional Scanning:
\begin{align}
    \mathbf{Y}_{\mathrm{fwd}} &= \text{SSM}_{\mathrm{scan}}(\mathbf{H}_{l-1}, \overline{\mathbf{A}}, \overline{\mathbf{B}}_{\mathrm{fwd}}, \mathbf{C}_{\mathrm{fwd}}) \\
    \mathbf{Y}_{\mathrm{bwd}} &= \text{SSM}_{\mathrm{scan}}(\text{Flip}(\mathbf{H}_{l-1}), \overline{\mathbf{A}}, \overline{\mathbf{B}}_{\mathrm{bwd}}, \mathbf{C}_{\mathrm{bwd}})
\end{align}

The final output is fused through a gating mechanism and mapped back to the time domain to obtain the vision branch output $\mathbf{H}_{\mathrm{vis}}$.
This architecture strictly limits the long-sequence modeling complexity to $O(N)$, achieving efficient capture of cross-period global dependencies in "temporal images" while significantly reducing memory usage.

\subsection{Adaptive Scale-Aware Fusion}
The three modalities provide complementary perspectives: the time-domain linear branch offers lossless numerical anchors, the frequency-domain wavelet branch parses periods and noise, and the vision Mamba branch captures long-range structural textures. Simple averaging is insufficient to address the dynamic shifts in importance among these modalities.

To address this, we design a Scale-Aware Gated Fusion module. A lightweight gating network generates time-step-level confidence weights:
\begin{equation}
    [\mathbf{G}_{\mathrm{time}}, \mathbf{G}_{\mathrm{freq}}, \mathbf{G}_{\mathrm{vis}}] = \text{Softmax}(\text{MLP}([\mathbf{H}_{\mathrm{time}}; \mathbf{H}_{\mathrm{freq}}; \mathbf{H}_{\mathrm{vis}}]))
\end{equation}

The final fused representation $\mathbf{H}_{\mathrm{fuse}}$ is computed as a weighted sum:
\begin{equation}
    \mathbf{H}_{\mathrm{fuse}} = \mathbf{G}_{\mathrm{time}} \odot \mathbf{H}_{\mathrm{time}} + \mathbf{G}_{\mathrm{freq}} \odot \mathbf{H}_{\mathrm{freq}} + \mathbf{G}_{\mathrm{vis}} \odot \mathbf{H}_{\mathrm{vis}}
\end{equation}

This mechanism ensures that the model flexibly leverages high-level features from the frequency and vision domains to refine predictions while maintaining the linear trends provided by $\mathbf{H}_{\mathrm{time}}$.

\section{Experiments}

\subsection{Experimental Setup}

\subsubsection{Datasets}
To comprehensively evaluate the performance of TriTS, we conduct experiments on seven widely used real-world benchmark datasets, covering diverse domains such as energy, transportation, and meteorology. Detailed statistics of these datasets are as follows:
\begin{enumerate}
    \item \textbf{ETT}: Comprises two years of data collected from power transformers, including ETTh1, ETTh2 (recorded hourly) and ETTm1, ETTm2 (recorded every 15 minutes)\cite{ett_data}.
    \item \textbf{Weather}: Contains 21 meteorological indicators recorded by the Max Planck Institute for Biogeochemistry meteorological station in Germany throughout 2020, sampled every 10 minutes\cite{autoformer}.
    \item \textbf{Electricity (ECL)}: Records the hourly power consumption data of 321 clients\cite{ecl_data}.
    \item \textbf{Traffic}: Sourced from 862 sensors on highways in the San Francisco Bay Area, recording hourly road occupancy rates\cite{ecl_data}.
\end{enumerate}

Detailed statistics of these datasets are presented in Table \ref{tab:data}. To explicitly evaluate the energy distribution among different components within the datasets, we employ SMA to decompose the time series. We then calculate the ratio of the covariance of the seasonal component to that of the trend component. This ratio serves as an indicator of the relative dominance of seasonality versus trend in the data. Additionally, the Forecastability metric\cite{forecastability} is utilized to quantify the regularity and predictability of the time series.

\begin{table}[htbp]
\centering
\caption{Dataset detailed descriptions. The dataset size is organized as (Train, Validation, Test).}
\label{tab:data}
\resizebox{\linewidth}{!}{
\begin{tabular}{lccccc}
\toprule
Dataset     & Dim  & Dataset Size           & Forecastability* & Information & $\frac{\text{cov(Season)}}{\text{cov(Trend)}}$ \\
\midrule
ETTm1       & 7    & $(34465, 11521, 11521)$ & 0.46             & Temperature & 0.073992 \\
ETTm2       & 7    & $(34465, 11521, 11521)$ & 0.55             & Temperature & 0.039469 \\
ETTh1       & 7    & $(8545, 2881, 2881)$    & 0.38             & Temperature & 0.490909 \\
ETTh2       & 7    & $(8545, 2881, 2881)$    & 0.45             & Temperature & 0.136543 \\
Electricity & 321  & $(18317, 2633, 5261)$   & 0.77             & Electricity & 11.836820 \\
Traffic     & 862  & $(12185, 1757, 3509)$   & 0.68             & Transportation      & 13.541667 \\
Weather     & 21   & $(36792, 5271, 10540)$  & 0.75             & Weather      & 0.005176 \\
\bottomrule
\end{tabular}
}
\end{table}

\subsubsection{Baseline Models}
To thoroughly assess the effectiveness of TriTS, we compare our method against 8 representative state-of-the-art (SOTA) models. These include: VisionTS\cite{visionts} and VisionTS++\cite{visionts++}, which are vision-based models adapted for time series forecasting via image pre-training; WPMixer\cite{wpmixer} and FDNet\cite{fdnet}, which are wavelet-based models; iTransformer\cite{itransformer} and PatchTST\cite{patchtst}, representing Transformer-based architectures; and TimeMixer\cite{timemixer} and DLinear\cite{dlinear}, representing linear models. These baselines span the vision, frequency, and time domains, providing a comprehensive benchmark against mainstream approaches.

\subsubsection{Experimental Configuration}
All experiments are implemented using the PyTorch framework on a server equipped with a single NVIDIA GeForce A800 80G GPU. The prediction length $T$ is set to $\{96, 192, 336, 720\}$. The Vim module of TriTS utilizes the Vim-Tiny architecture, pre-trained on ImageNet-1k. For the modality-specific hyperparameters, the multi-level wavelet decomposition is fixed at $m=3$ levels. Additionally, the patch size is empirically tuned from the candidate set $\{8, 12, 16\}$ based on the specific characteristics of each dataset. We uniformly employ the Adam optimizer with a batch size of 128. Model training incorporates an Early Stopping strategy with a patience of 5 epochs. The evaluation metric is Mean Squared Error (MSE). To ensure a fair evaluation, the implementations of baseline models follow their official codebases, and the reported results are comprehensively aligned with the standardized benchmarks established by the time series community. Prior to network input, all data undergo standardization via Zero-Mean Normalization.

\subsubsection{Comparative Experimental Results}
Tables \ref{tab:results} and \ref{tab:comparison_sota} present the comparative results of TriTS and various baseline models on multivariate long-term time series forecasting tasks. Lower MSE and MAE values indicate superior performance.

\begin{table*}[htbp]
\centering
\caption{Performance comparison of long-term time series forecasting across different datasets with lookback window=96. The best results are highlighted in \textbf{\underline{bold and underlined}}.}
\label{tab:results}
\resizebox{\linewidth}{!}{ 
\setlength{\tabcolsep}{3.5pt} 
\begin{tabular}{llcccccccccccccccccc}
\toprule
\multirow{2}{*}{\textbf{Dataset}} & \multirow{2}{*}{\textbf{Len}} & \multicolumn{2}{c}{\textbf{TriTS}} & \multicolumn{2}{c}{VisionTS++} & \multicolumn{2}{c}{VisionTS} & \multicolumn{2}{c}{FDNet} & \multicolumn{2}{c}{WPMixer} & \multicolumn{2}{c}{iTransformer} & \multicolumn{2}{c}{PatchTST} & \multicolumn{2}{c}{TimeMixer} & \multicolumn{2}{c}{DLinear} \\
\cmidrule(lr){3-4} \cmidrule(lr){5-6} \cmidrule(lr){7-8} \cmidrule(lr){9-10} \cmidrule(lr){11-12} \cmidrule(lr){13-14} \cmidrule(lr){15-16} \cmidrule(lr){17-18} \cmidrule(lr){19-20}
 &  & MSE & MAE & MSE & MAE & MSE & MAE & MSE & MAE & MSE & MAE & MSE & MAE & MSE & MAE & MSE & MAE & MSE & MAE \\ 
\midrule

\multirow{2}{*}{ETTh1} 
 & 720 & \textbf{\underline{0.449}} & \textbf{\underline{0.440}} & 0.456 & 0.447 & 0.455 & 0.442 & 0.470 & 0.465 & 0.464 & 0.458 & 0.509 & 0.494 & 0.488 & 0.477 & 0.498 & 0.482 & 0.513 & 0.510 \\
 & AVG & \textbf{\underline{0.411}} & \textbf{\underline{0.425}} & 0.423 & 0.437 & 0.432 & 0.443 & 0.435 & 0.431 & 0.433 & 0.430 & 0.457 & 0.447 & 0.455 & 0.444 & 0.447 & 0.440 & 0.461 & 0.457 \\ 
\midrule

\multirow{2}{*}{ETTh2} 
 & 720 & \textbf{\underline{0.401}} & \textbf{\underline{0.421}} & 0.421 & 0.439 & 0.425 & 0.441 & 0.425 & 0.442 & 0.426 & 0.444 & 0.426 & 0.445 & 0.436 & 0.453 & 0.412 & 0.434 & 0.839 & 0.661 \\
 & AVG & 0.366 & \textbf{\underline{0.380}} & 0.379 & 0.402 & 0.397 & 0.421 & 0.373 & 0.399 & 0.378 & 0.405 & 0.383 & 0.407 & 0.385 & 0.410 & \textbf{\underline{0.364}} & 0.395 & 0.563 & 0.519 \\ 
\midrule

\multirow{2}{*}{ETTm1} 
 & 720 & 0.447 & 0.428 & 0.463 & 0.448 & 0.461 & 0.442 & 0.465 & 0.440 & 0.450 & 0.441 & 0.486 & 0.456 & 0.475 & 0.453 & \textbf{\underline{0.390}} & \textbf{\underline{0.404}} & 0.415 & 0.415 \\
 & AVG & \textbf{\underline{0.379}} & \textbf{\underline{0.386}} & 0.392 & 0.398 & 0.401 & 0.405 & 0.385 & 0.394 & 0.380 & 0.396 & 0.407 & 0.412 & 0.396 & 0.406 & 0.381 & 0.395 & 0.404 & 0.408 \\ 
\midrule

\multirow{2}{*}{ETTm2} 
 & 720 & \textbf{\underline{0.361}} & \textbf{\underline{0.365}} & 0.381 & 0.388 & 0.384 & 0.392 & 0.399 & 0.397 & 0.401 & 0.399 & 0.412 & 0.406 & 0.407 & 0.401 & 0.391 & 0.396 & 0.558 & 0.525 \\
 & AVG & \textbf{\underline{0.270}} & \textbf{\underline{0.318}} & 0.294 & 0.327 & 0.297 & 0.331 & 0.279 & 0.324 & 0.279 & 0.325 & 0.291 & 0.334 & 0.284 & 0.327 & 0.275 & 0.323 & 0.354 & 0.402 \\ 
\midrule

\multirow{2}{*}{Weather} 
 & 720 & \textbf{\underline{0.318}} & \textbf{\underline{0.323}} & 0.332 & 0.341 & 0.336 & 0.348 & 0.346 & 0.343 & 0.346 & 0.346 & 0.361 & 0.351 & 0.365 & 0.367 & 0.339 & 0.341 & 0.345 & 0.382 \\
 & AVG & \textbf{\underline{0.221}} & \textbf{\underline{0.238}} & 0.224 & 0.243 & 0.229 & 0.248 & 0.246 & 0.271 & 0.245 & 0.274 & 0.261 & 0.281 & 0.261 & 0.285 & 0.240 & 0.271 & 0.265 & 0.315 \\ 
\midrule

\multirow{2}{*}{ECL} 
 & 720 & \textbf{\underline{0.188}} & \textbf{\underline{0.279}} & 0.194 & 0.283 & 0.191 & \textbf{\underline{0.279}} & 0.192 & 0.287 & 0.224 & 0.311 & 0.228 & 0.312 & 0.230 & 0.311 & 0.225 & 0.310 & 0.258 & 0.350 \\
 & AVG & \textbf{\underline{0.158}} & 0.260 & 0.164 & 0.267 & 0.161 & 0.263 & 0.162 & \textbf{\underline{0.257}} & 0.181 & 0.271 & 0.180 & 0.270 & 0.190 & 0.275 & 0.182 & 0.272 & 0.225 & 0.315 \\ 
\midrule

\multirow{2}{*}{Traffic} 
 & 720 & \textbf{\underline{0.441}} & \textbf{\underline{0.270}} & 0.447 & 0.276 & 0.446 & 0.273 & 0.451 & 0.291 & 0.529 & 0.324 & 0.461 & 0.302 & 0.500 & 0.309 & 0.506 & 0.313 & 0.645 & 0.394 \\
 & AVG & \textbf{\underline{0.399}} & \textbf{\underline{0.258}} & 0.413 & 0.269 & 0.413 & 0.272 & 0.408 & 0.266 & 0.486 & 0.306 & 0.423 & 0.283 & 0.467 & 0.293 & 0.484 & 0.297 & 0.625 & 0.383 \\ 
\bottomrule
\end{tabular}
}
\end{table*}

\begin{table*}[htbp]
\centering
\caption{Performance comparison of long-term time series forecasting across diverse datasets with hyperparameter tuning. The best results are highlighted in \textbf{\underline{bold and underlined}}.}
\label{tab:comparison_sota}
\resizebox{\textwidth}{!}{%
\setlength{\tabcolsep}{3pt}
\begin{tabular}{lcccccccccccccc}
\toprule
\multirow{2}{*}{\textbf{Dataset}} & \multicolumn{2}{c}{\textbf{TriTS}} & \multicolumn{2}{c}{VisionTS} & \multicolumn{2}{c}{VisionTS++} & \multicolumn{2}{c}{WPMixer} & \multicolumn{2}{c}{iTransformer} & \multicolumn{2}{c}{PatchTST} & \multicolumn{2}{c}{TimeMixer} \\
\cmidrule(lr){2-3} \cmidrule(lr){4-5} \cmidrule(lr){6-7} \cmidrule(lr){8-9} \cmidrule(lr){10-11} \cmidrule(lr){12-13} \cmidrule(lr){14-15}
 & MSE & MAE & MSE & MAE & MSE & MAE & MSE & MAE & MSE & MAE & MSE & MAE & MSE & MAE \\ 
\midrule

Weather & \textbf{\underline{0.218}} & 0.250 & 0.269 & 0.292 & 0.226 & \textbf{\underline{0.243}} & 0.220 & 0.255 & 0.232 & 0.270 & 0.241 & 0.264 & 0.222 & 0.262 \\
ECL & 0.165 & 0.252 & 0.207 & 0.294 & 0.181 & 0.264 & 0.159 & 0.251 & 0.163 & 0.258 & 0.159 & 0.253 & \textbf{\underline{0.156}} & \textbf{\underline{0.247}} \\
ETTh1 & \textbf{\underline{0.385}} & \textbf{\underline{0.412}} & 0.390 & 0.414 & 0.403 & 0.418 & 0.399 & 0.424 & 0.439 & 0.448 & 0.413 & 0.434 & 0.411 & 0.423 \\
ETTh2 & \textbf{\underline{0.312}} & 0.372 & 0.333 & 0.375 & 0.327 & \textbf{\underline{0.365}} & 0.328 & 0.379 & 0.370 & 0.403 & 0.324 & 0.381 & 0.316 & 0.384 \\
ETTm1 & \textbf{\underline{0.331}} & \textbf{\underline{0.359}} & 0.374 & 0.372 & 0.354 & 0.369 & 0.348 & 0.378 & 0.361 & 0.390 & 0.353 & 0.382 & 0.348 & 0.375 \\
ETTm2 & 0.249 & 0.311 & 0.282 & 0.321 & \textbf{\underline{0.244}} & \textbf{\underline{0.298}} & 0.256 & 0.315 & 0.269 & 0.327 & 0.256 & 0.317 & 0.256 & 0.315 \\ 

\bottomrule
\end{tabular}%
}
\end{table*}

Benefiting from the comprehensive data capture across the tri-modal branches, TriTS achieves optimal performance on the majority of datasets and prediction lengths. Compared to VisionTS and VisionTS++, which rely solely on the vision modality, TriTS effectively compensates for the potential lack of numerical fidelity in vision models by integrating the time-domain streaming branch and the frequency-domain wavelet branch. This is particularly evident in the ETT datasets, which exhibit substantial numerical fluctuations. Furthermore, although WPMixer and FDNet employ multi-resolution wavelet mixing, they lack the capability for global context visual perception. The Visual Mamba branch in TriTS addresses this deficiency, resulting in superior performance in long-sequence forecasting (e.g., $L=720$).

\subsection{Ablation Experiments}
To verify the contribution of each core component within the TriTS framework, we conduct ablation studies. We re-evaluated the model's performance on selected datasets after removing each specific module. The results are detailed in Table \ref{tab:ablation}.

\begin{table*}[htbp]
    \centering
    \caption{Ablation study of different components in TriTS. The best results are highlighted in \textbf{bold}.}
    \label{tab:ablation}
    \resizebox{\textwidth}{!}{
    \begin{tabular}{lcccccccc}
        \toprule
        \multirow{2}{*}{\textbf{Model Variants}} & \multicolumn{2}{c}{\textbf{ETTh1}} & \multicolumn{2}{c}{\textbf{ETTm1}} & \multicolumn{2}{c}{\textbf{Weather}} & \multicolumn{2}{c}{\textbf{ECL}} \\
        \cmidrule(lr){2-3} \cmidrule(lr){4-5} \cmidrule(lr){6-7} \cmidrule(lr){8-9}
         & MSE & MAE & MSE & MAE & MSE & MAE & MSE & MAE \\
        \midrule
        \textbf{TriTS (Full Model)} & \textbf{0.411} & \textbf{0.425} & \textbf{0.379} & \textbf{0.386} & \textbf{0.221} & \textbf{0.238} & \textbf{0.158} & \textbf{0.260} \\
        \midrule
        w/o Time Branch & 1.213 & 1.552 & 0.558 & 0.648 & 0.281 & 0.262 & 0.829 & 1.211 \\
        w/o Freq Branch & 0.447 & 0.481 & 0.402 & 0.415 & 0.301 & 0.328 & 0.196 & 0.320 \\
        w/o Vision Branch & 0.474 & 0.499 & 0.423 & 0.449 & 0.251 & 0.288 & 0.164 & 0.271 \\
        w/o Gating Fusion & 0.442 & 0.439 & 0.388 & 0.391 & 0.224 & 0.244 & 0.166 & 0.270 \\
        \bottomrule
    \end{tabular}
    }
\end{table*}

As demonstrated in Table \ref{tab:ablation}, the removal of any module leads to a degradation in performance. Notably, eliminating the time-domain branch causes the model to exhibit severe instability or gradient overfitting during training. This underscores the role of the time domain as the foundational backbone of the entire forecasting process, ensuring numerical stability.

\begin{figure}[htbp]
    \centering
    \begin{subfigure}[b]{0.48\textwidth}
        \centering
        \includegraphics[width=\linewidth]{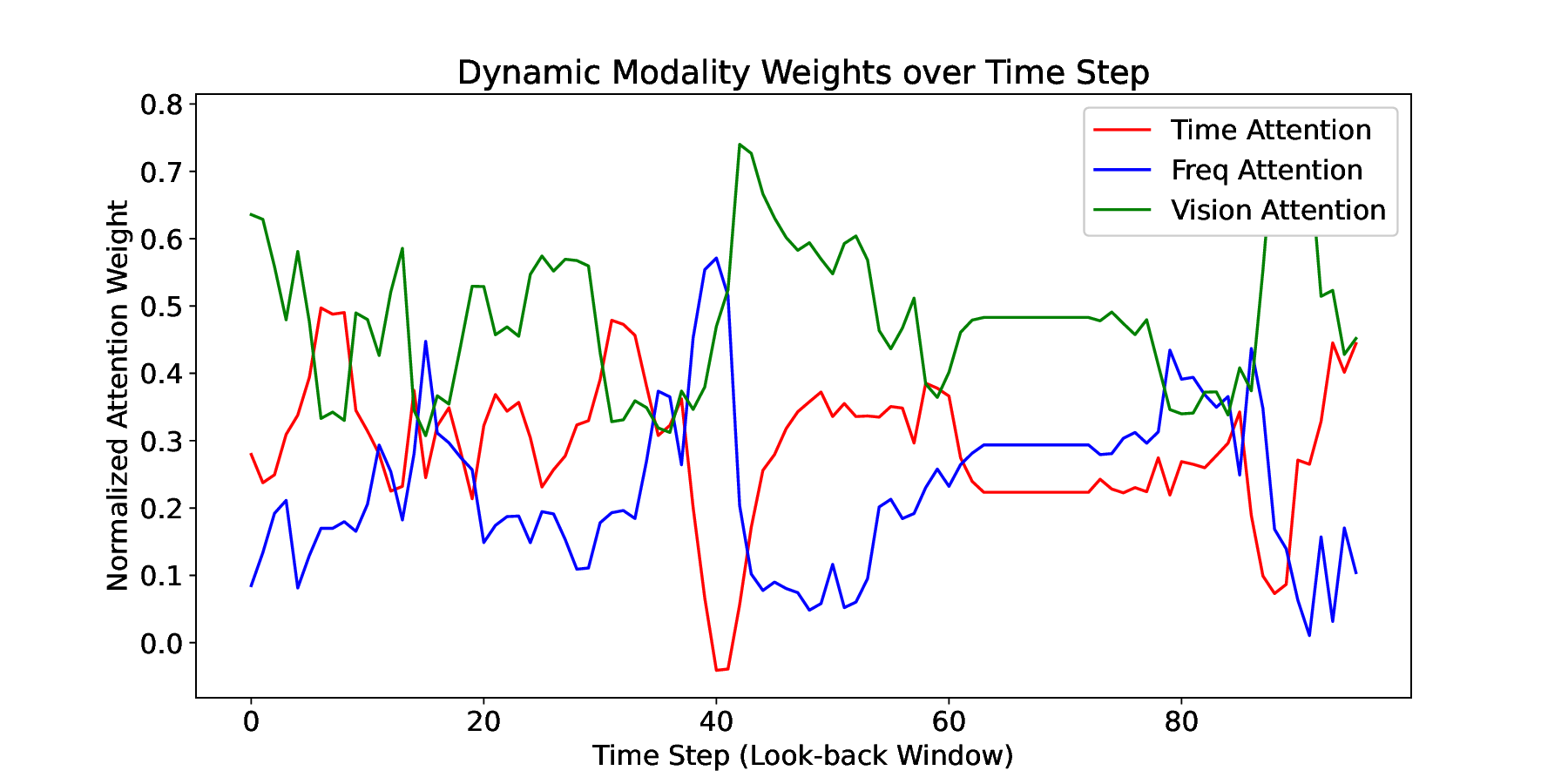}
        \caption{ETTh1 dataset}
        \label{fig:weight_etth1}
    \end{subfigure}
    \hfill 
    \begin{subfigure}[b]{0.48\textwidth}
        \centering
        \includegraphics[width=\linewidth]{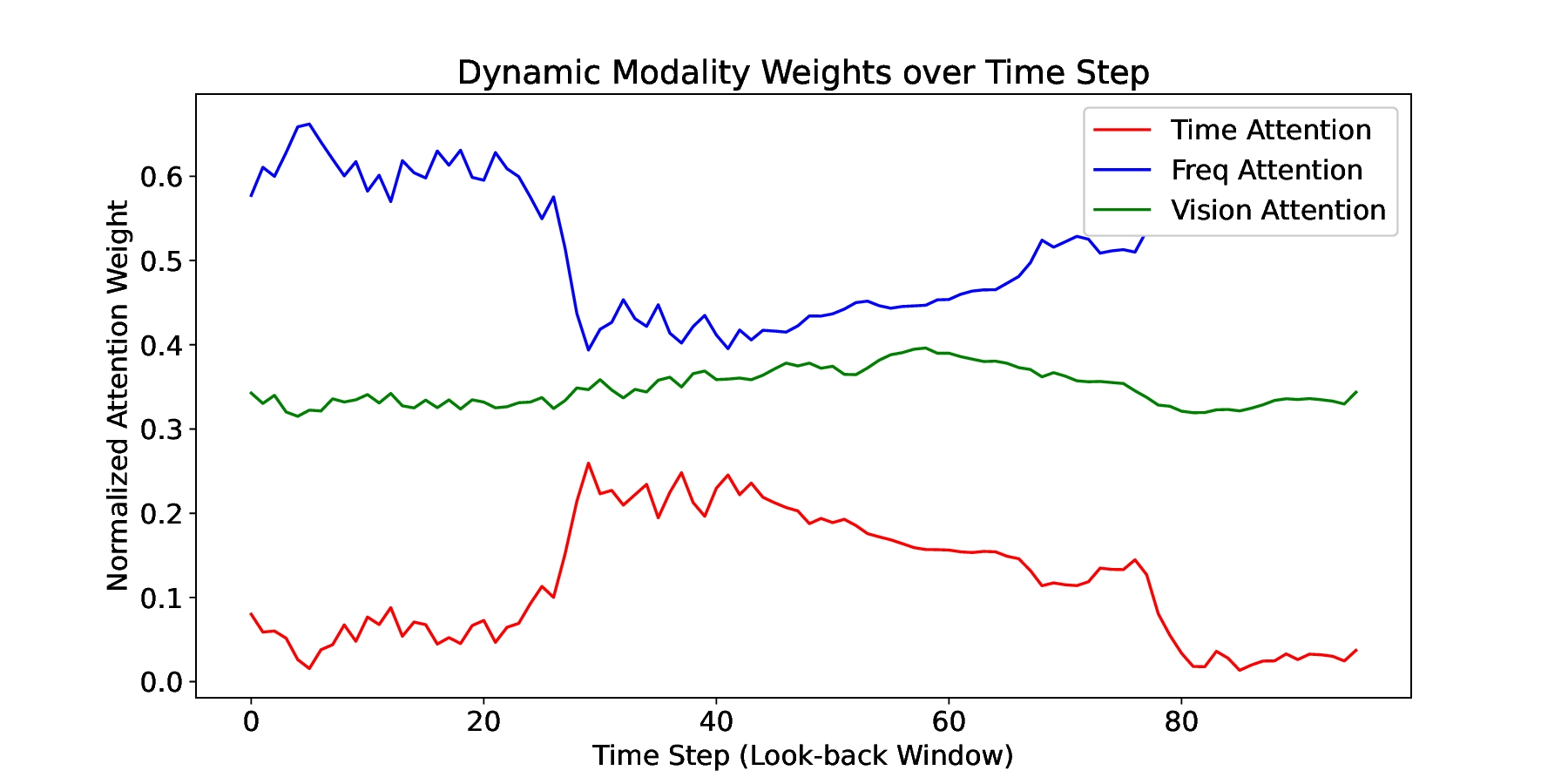}
        \caption{Weather dataset}
        \label{fig:weight_weather}
    \end{subfigure}
    
    \caption{Visualization of modality weights on different datasets. (a) shows the weights on ETTh1, reflecting strong long-term dependencies; (b) shows the weights on Weather, indicating stronger local periodic patterns.}
    \label{fig:weight_visualization_group}
\end{figure}

\subsection{Further Analysis}
To further analyze the contribution of each modality to the final prediction, we visualize the fusion weights of the modules, as shown in Figure \ref{fig:weight_visualization_group}.

On the ETTh1 dataset, the weight assigned to the vision domain is notably higher than that of other domains, reflecting the presence of strong long-term global dependencies in the data. Conversely, on the Weather dataset, the weights for the vision and time domains are lower, while the frequency domain weight is elevated. This indicates that the Weather dataset is characterized by stronger local periodic relationships, necessitating enhanced frequency-domain modeling to improve predictive capability.

\subsection{Model Efficiency Analysis}
\begin{figure}[htbp]
    \centering
    \includegraphics[width=0.8\linewidth]{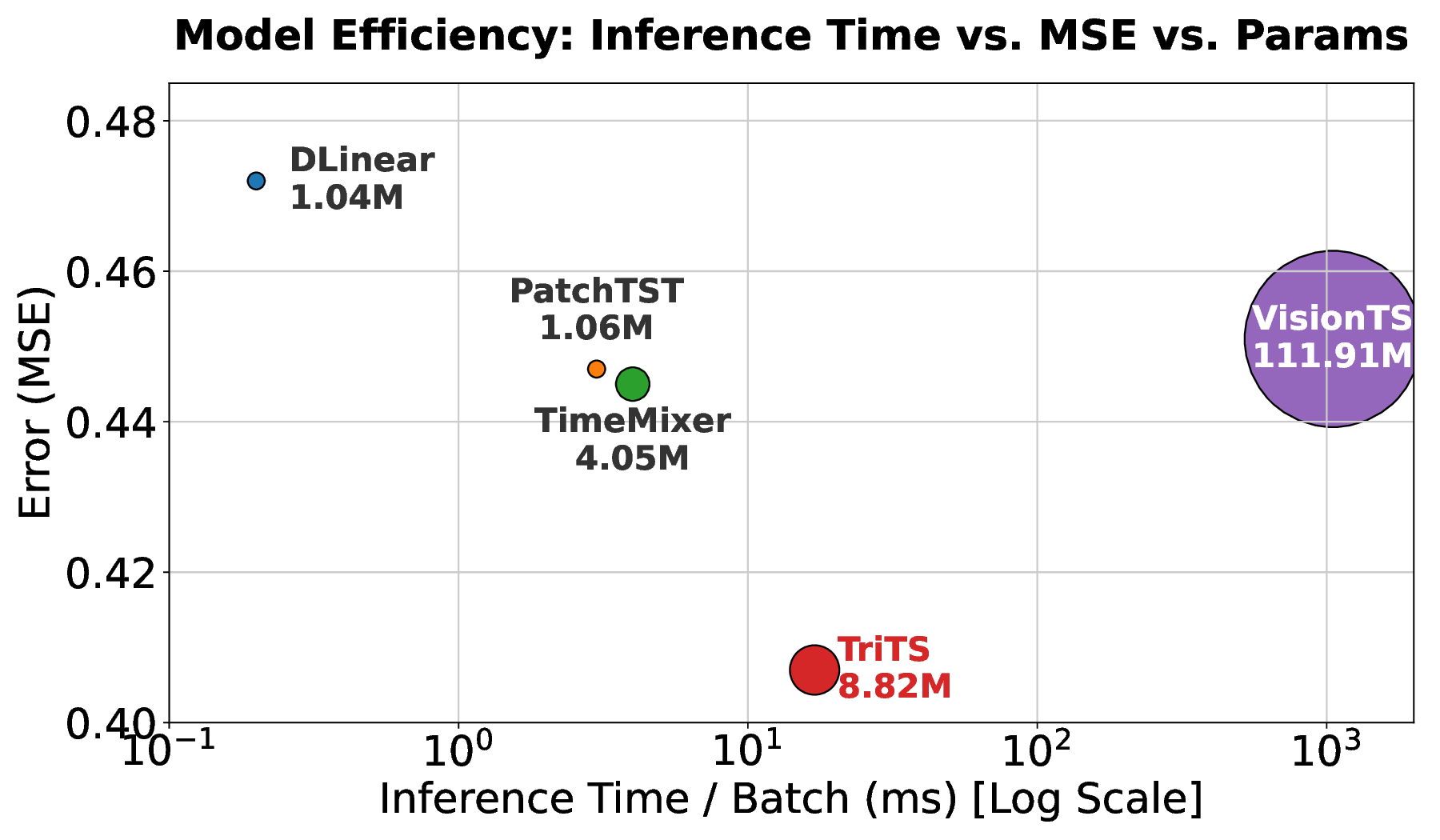}
    \caption{Computational efficiency comparison between TriTS and baseline models. All evaluations are conducted on the ETTh1 dataset with both look-back and prediction windows set to 720.}
    \label{fig:efficiency}
\end{figure}

In addition to prediction accuracy, we also evaluate the model's operational efficiency. Figure \ref{fig:efficiency} compares TriTS with VisionTS, PatchTST, and TimeMixer in terms of training time and memory usage. By adopting a multimodal framework, TriTS eliminates the need for large-scale vision models; the Vim-Tiny backbone is sufficient to achieve high performance, whereas VisionTS relies on the significantly larger MAE-Base model. Furthermore, by replacing the quadratic-complexity ViT with the linear-complexity Visual Mamba, and combining it with efficient linear and wavelet modules, TriTS achieves significantly lower memory usage than VisionTS when processing long sequences. Consequently, TriTS proves to be more compact and efficient than single-modal methods while delivering distinct performance improvements over VisionTS.

\section{Conclusion}
We propose a multimodal time series forecasting framework that integrates time, frequency, and vision domains, leveraging the complementary strengths of each modality to enhance forecasting capability. Experiments demonstrate that TriTS achieves state-of-the-art performance across multiple benchmark datasets while significantly reducing computational overhead and memory usage compared to existing methods. The core innovations of this paper, including bidirectional selective state space modeling for the vision branch and multi-resolution wavelet mixing for the frequency branch, offer novel perspectives for addressing long-term time series forecasting tasks. Future work will explore adaptive period detection methods to further enhance the flexibility of the vision branch and extend the framework to more complex scenarios, such as multivariate irregular time series forecasting.

\section*{Acknowledgment}
The authors would like to express their sincere gratitude to Professor Siyang Lu and Professor Jing Wang for their invaluable guidance and support throughout this research. We also thank Dr. Xiaoyu Ou for his insightful discussions and technical assistance. Furthermore, this work was supported by the School of Computer and Information Technology at Beijing Jiaotong University.

\section*{Declaration of Generative AI in the Writing Process}
During the preparation of this work, the authors used Gemini 3 in order to improve the language clarity and refine the English expression of the manuscript. After using this tool, the authors reviewed and edited the content as needed and take full responsibility for the content of the publication.


{
    \small
    \bibliographystyle{ieeenat_fullname}
    \bibliography{main}

@INPROCEEDINGS{energe,
  author={Sun, Qianxiang and Ma, Hongyuan and Li, Guangdi and Li, Ziwen and Wang, Yining},
  booktitle={2022 First International Conference on Cyber-Energy Systems and Intelligent Energy (ICCSIE)}, 
  title={Short-Term Multivariate Load Forecasting for Integrated Energy Systems Based on BIGRU-AM and Multi-Task Learning}, 
  year={2023},
  volume={},
  number={},
  pages={1-6},
  doi={10.1109/ICCSIE55183.2023.10175297}}

@ARTICLE{traffic,
  author={Hu, Yang and Li, Shaobo and Xia, Dawen and Zhang, Wenyong and Yuan, Panliang and Wu, Fengbin and Li, Huaqing},
  journal={IEEE Internet of Things Journal}, 
  title={A Multiview Spatial-Temporal Adaptive Transformer-GRU Framework for Traffic Flow Prediction}, 
  year={2025},
  volume={12},
  number={6},
  pages={7114-7132},
  doi={10.1109/JIOT.2024.3496795}}

@inproceedings{finance,
author = {Ao, Xiang},
title = {Research on the Stock Price Prediction Model Based on Large-Scale Transactions},
year = {2025},
isbn = {9798400718748},
publisher = {Association for Computing Machinery},
address = {New York, NY, USA},
url = {https://doi.org/10.1145/3773365.3773427},
doi = {10.1145/3773365.3773427},
booktitle = {Proceedings of the 2025 8th International Conference on Computer Information Science and Artificial Intelligence},
pages = {392–396},
numpages = {5},
location = {
},
series = {CISAI '25}
}

@article{weather,
  title={Interpretable weather forecasting for worldwide stations with a unified deep model},
  author={Wu, Haixu and Zhou, Hang and Long, Mingsheng and Wang, Jianmin},
  journal={Nature Machine Intelligence},
  volume={5},
  number={6},
  pages={602--611},
  year={2023},
  publisher={Nature Publishing Group},
  doi={10.1038/s42256-023-00667-9},
  url={https://doi.org/10.1038/s42256-023-00667-9}
}

@article{industrial,
  author={Nizam, Hussain and Zafar, Samra and Lv, Zefeng and Wang, Fan and Hu, Xiaopeng},
  journal={IEEE Sensors Journal}, 
  title={Real-Time Deep Anomaly Detection Framework for Multivariate Time-Series Data in Industrial IoT}, 
  year={2022},
  volume={22},
  number={23},
  pages={22836-22849},
  doi={10.1109/JSEN.2022.3211874}}

@article{lstm,
author = {Hochreiter, Sepp and Schmidhuber, J\"{u}rgen},
title = {Long Short-Term Memory},
year = {1997},
issue_date = {November 15, 1997},
publisher = {MIT Press},
address = {Cambridge, MA, USA},
volume = {9},
number = {8},
issn = {0899-7667},
url = {https://doi.org/10.1162/neco.1997.9.8.1735},
doi = {10.1162/neco.1997.9.8.1735},
journal = {Neural Comput.},
month = nov,
pages = {1735–1780},
numpages = {46}
}

@article{rnn,
  title={Learning representations by back-propagating errors},
  author={Rumelhart, David E. and Hinton, Geoffrey E. and Williams, Ronald J.},
  journal={Nature},
  volume={323},
  number={6088},
  pages={533--536},
  year={1986},
  month={oct},
  publisher={Nature Publishing Group},
  doi={10.1038/323533a0},
  url={https://doi.org/10.1038/323533a0}
}

@inproceedings{attention,
 author = {Vaswani, Ashish and Shazeer, Noam and Parmar, Niki and Uszkoreit, Jakob and Jones, Llion and Gomez, Aidan N and Kaiser, \L ukasz and Polosukhin, Illia},
 booktitle = {Advances in Neural Information Processing Systems},
 editor = {I. Guyon and U. Von Luxburg and S. Bengio and H. Wallach and R. Fergus and S. Vishwanathan and R. Garnett},
 pages = {},
 publisher = {Curran Associates, Inc.},
 title = {Attention is All you Need},
 url = {https://proceedings.neurips.cc/paper_files/paper/2017/file/3f5ee243547dee91fbd053c1c4a845aa-Paper.pdf},
 volume = {30},
 year = {2017}
}

@misc{patchtst,
      title={A Time Series is Worth 64 Words: Long-term Forecasting with Transformers}, 
      author={Yuqi Nie and Nam H. Nguyen and Phanwadee Sinthong and Jayant Kalagnanam},
      year={2023},
      eprint={2211.14730},
      archivePrefix={arXiv},
      primaryClass={cs.LG},
      url={https://arxiv.org/abs/2211.14730}, 
}

@misc{autoformer,
      title={Autoformer: Decomposition Transformers with Auto-Correlation for Long-Term Series Forecasting}, 
      author={Haixu Wu and Jiehui Xu and Jianmin Wang and Mingsheng Long},
      year={2022},
      eprint={2106.13008},
      archivePrefix={arXiv},
      primaryClass={cs.LG},
      url={https://arxiv.org/abs/2106.13008}, 
}

@misc{fedformer,
      title={FEDformer: Frequency Enhanced Decomposed Transformer for Long-term Series Forecasting}, 
      author={Tian Zhou and Ziqing Ma and Qingsong Wen and Xue Wang and Liang Sun and Rong Jin},
      year={2022},
      eprint={2201.12740},
      archivePrefix={arXiv},
      primaryClass={cs.LG},
      url={https://arxiv.org/abs/2201.12740}, 
}

@misc{filternet,
      title={FilterNet: Harnessing Frequency Filters for Time Series Forecasting}, 
      author={Kun Yi and Jingru Fei and Qi Zhang and Hui He and Shufeng Hao and Defu Lian and Wei Fan},
      year={2024},
      eprint={2411.01623},
      archivePrefix={arXiv},
      primaryClass={cs.LG},
      url={https://arxiv.org/abs/2411.01623}, 
}

@misc{visionts,
title={Vision{TS}: Visual Masked Autoencoders Are Free-Lunch Zero-Shot Time Series Forecasters},
author={Mouxiang Chen and Lefei Shen and Zhuo Li and Xiaoyun Joy Wang and Jianling Sun and Chenghao Liu},
year={2025},
url={https://openreview.net/forum?id=IEs29RYxfK}
}

@misc{visionts++,
      title={VisionTS++: Cross-Modal Time Series Foundation Model with Continual Pre-trained Vision Backbones}, 
      author={Lefei Shen and Mouxiang Chen and Xu Liu and Han Fu and Xiaoxue Ren and Jianling Sun and Zhuo Li and Chenghao Liu},
      year={2025},
      eprint={2508.04379},
      archivePrefix={arXiv},
      primaryClass={cs.CV},
      url={https://arxiv.org/abs/2508.04379}, 
}

@misc{mae,
      title={Masked Autoencoders Are Scalable Vision Learners}, 
      author={Kaiming He and Xinlei Chen and Saining Xie and Yanghao Li and Piotr Dollár and Ross Girshick},
      year={2021},
      eprint={2111.06377},
      archivePrefix={arXiv},
      primaryClass={cs.CV},
      url={https://arxiv.org/abs/2111.06377}, 
}

@inproceedings{logtrans,
  author={Nie, Xingqing and Zhou, Xiaogen and Li, Zhiqiang and Wang, Luoyan and Lin, Xingtao and Tong, Tong},
  booktitle={2022 IEEE 24th Int Conf on High Performance Computing}, 
  title={LogTrans: Providing Efficient Local-Global Fusion with Transformer and CNN Parallel Network for Biomedical Image Segmentation}, 
  year={2022},
  volume={},
  number={},
  pages={769-776},
  doi={10.1109/HPCC-DSS-SmartCity-DependSys57074.2022.00128}}

@article{manba,
  title={Mamba: Linear-Time Sequence Modeling with Selective State Spaces},
  author={Albert Gu and Tri Dao},
  journal={ArXiv},
  year={2023},
  volume={abs/2312.00752},
  url={https://api.semanticscholar.org/CorpusID:265551773}
}

@inproceedings{xpatch,
  title={xPatch: Dual-Stream Time Series Forecasting with Exponential Seasonal-Trend Decomposition},
  author={Stitsyuk, A. and Choi, J.},
  booktitle={Proceedings of the AAAI Conference on Artificial Intelligence},
  volume={39},
  number={19},
  pages={20601--20609},
  year={2025},
  doi={10.1609/aaai.v39i19.34270},
  url={https://doi.org/10.1609/aaai.v39i19.34270}
}

@inproceedings{vim,
author = {Zhu, Lianghui and Liao, Bencheng and Zhang, Qian and Wang, Xinlong and Liu, Wenyu and Wang, Xinggang},
title = {Vision mamba: efficient visual representation learning with bidirectional state space model},
year = {2024},
publisher = {JMLR.org},
booktitle = {Proceedings of the 41st International Conference on Machine Learning},
articleno = {2584},
numpages = {14},
location = {Vienna, Austria},
series = {ICML'24}
}

@misc{itransformer,
      title={iTransformer: Inverted Transformers Are Effective for Time Series Forecasting}, 
      author={Yong Liu and Tengge Hu and Haoran Zhang and Haixu Wu and Shiyu Wang and Lintao Ma and Mingsheng Long},
      year={2024},
      eprint={2310.06625},
      archivePrefix={arXiv},
      primaryClass={cs.LG},
      url={https://arxiv.org/abs/2310.06625}, 
}

@inproceedings{timemixer,
 author = {Wang, Shiyu and Wu, Haixu and Shi, Xiaoming and Hu, Tengge and Luo, Huakun and Ma, Lintao and Zhang, James and ZHOU, JUN},
 booktitle = {International Conference on Representation Learning},
 editor = {B. Kim and Y. Yue and S. Chaudhuri and K. Fragkiadaki and M. Khan and Y. Sun},
 pages = {38626--38652},
 title = {TimeMixer: Decomposable Multiscale Mixing for Time Series Forecasting},
 url = {https://proceedings.iclr.cc/paper_files/paper/2024/file/a7ac8a21e5a27e7ab31a5f42a0117bdb-Paper-Conference.pdf},
 volume = {2024},
 year = {2024}
}

@inproceedings{wpmixer,
author = {Murad, Md Mahmuddun Nabi and Aktukmak, Mehmet and Yilmaz, Yasin},
title = {WPMixer: efficient multi-resolution mixing for long-term time series forecasting},
year = {2025},
isbn = {978-1-57735-897-8},
publisher = {AAAI Press},
url = {https://doi.org/10.1609/aaai.v39i18.34156},
doi = {10.1609/aaai.v39i18.34156},
booktitle = {Proceedings of the Thirty-Ninth AAAI Conference on Artificial Intelligence and Thirty-Seventh Conference on Innovative Applications of Artificial Intelligence and Fifteenth Symposium on Educational Advances in Artificial Intelligence},
articleno = {2183},
numpages = {8},
series = {AAAI'25/IAAI'25/EAAI'25}
}

@article{fdnet,
title = {FDNet: High-frequency disentanglement network with information-theoretic guidance for multivariate time series forecasting},
journal = {Pattern Recognition},
volume = {173},
pages = {112810},
year = {2026},
issn = {0031-3203},
doi = {https://doi.org/10.1016/j.patcog.2025.112810},
url = {https://www.sciencedirect.com/science/article/pii/S0031320325014736},
author = {Ao Hu and Liangjian Wen and Jiang Duan and Yong Dai and Dongkai Wang and Shudong Huang and Jun Wang and Zenglin Xu}
}

@inproceedings{yolo,
  author={Redmon, Joseph and Divvala, Santosh and Girshick, Ross and Farhadi, Ali},
  booktitle={2016 IEEE Conference on Computer Vision and Pattern Recognition (CVPR)}, 
  title={You Only Look Once: Unified, Real-Time Object Detection}, 
  year={2016},
  volume={},
  number={},
  pages={779-788},
  doi={10.1109/CVPR.2016.91}}

@inproceedings{resnet,
  author={He, Kaiming and Zhang, Xiangyu and Ren, Shaoqing and Sun, Jian},
  booktitle={2016 IEEE Conference on Computer Vision and Pattern Recognition (CVPR)}, 
  title={Deep Residual Learning for Image Recognition}, 
  year={2016},
  volume={},
  number={},
  pages={770-778},
  doi={10.1109/CVPR.2016.90}}

@inproceedings{vit,
title={An Image is Worth 16x16 Words: Transformers for Image Recognition at Scale},
author={Alexey Dosovitskiy and Lucas Beyer and Alexander Kolesnikov and Dirk Weissenborn and Xiaohua Zhai and Thomas Unterthiner and Mostafa Dehghani and Matthias Minderer and Georg Heigold and Sylvain Gelly and Jakob Uszkoreit and Neil Houlsby},
booktitle={International Conference on Learning Representations},
year={2021},
url={https://openreview.net/forum?id=YicbFdNTTy}
}

@misc{dlinear,
      title={Are Transformers Effective for Time Series Forecasting?}, 
      author={Ailing Zeng and Muxi Chen and Lei Zhang and Qiang Xu},
      year={2022},
      eprint={2205.13504},
      archivePrefix={arXiv},
      primaryClass={cs.AI},
      url={https://arxiv.org/abs/2205.13504}, 
}

@inproceedings{timesnet,
title={TimesNet: Temporal 2D-Variation Modeling for General Time Series Analysis},
author={Haixu Wu and Tengge Hu and Yong Liu and Hang Zhou and Jianmin Wang and Mingsheng Long},
booktitle={The Eleventh International Conference on Learning Representations },
year={2023},
url={https://openreview.net/forum?id=ju_Uqw384Oq}
}

@misc{ett_data,
      title={Informer: Beyond Efficient Transformer for Long Sequence Time-Series Forecasting}, 
      author={Haoyi Zhou and Shanghang Zhang and Jieqi Peng and Shuai Zhang and Jianxin Li and Hui Xiong and Wancai Zhang},
      year={2021},
      eprint={2012.07436},
      archivePrefix={arXiv},
      primaryClass={cs.LG},
      url={https://arxiv.org/abs/2012.07436}, 
}

@inproceedings{ecl_data,
author = {Lai, Guokun and Chang, Wei-Cheng and Yang, Yiming and Liu, Hanxiao},
title = {Modeling Long- and Short-Term Temporal Patterns with Deep Neural Networks},
year = {2018},
isbn = {9781450356572},
publisher = {Association for Computing Machinery},
address = {New York, NY, USA},
url = {https://doi.org/10.1145/3209978.3210006},
doi = {10.1145/3209978.3210006},
booktitle = {The 41st International ACM SIGIR Conference on Research \& Development in Information Retrieval},
pages = {95–104},
numpages = {10},
location = {Ann Arbor, MI, USA},
series = {SIGIR '18}
}

@inproceedings{forecastability,
  title = 	 {Forecastable Component Analysis},
  author = 	 {Goerg, Georg},
  booktitle = 	 {Proceedings of the 30th International Conference on Machine Learning},
  pages = 	 {64--72},
  year = 	 {2013},
  editor = 	 {Dasgupta, Sanjoy and McAllester, David},
  volume = 	 {28},
  number =       {2},
  series = 	 {Proceedings of Machine Learning Research},
  address = 	 {Atlanta, Georgia, USA},
  month = 	 {17--19 Jun},
  publisher =    {PMLR},
  url = 	 {https://proceedings.mlr.press/v28/goerg13.html}
}
}


\end{document}